\documentclass{article}

\PassOptionsToPackage{numbers, compress}{natbib}

\usepackage[final]{neurips_2024}

\usepackage{xurl}            %

\usepackage{booktabs}       %
\usepackage{amsfonts}       %
\usepackage{nicefrac}       %
\usepackage{microtype}      %
\usepackage[dvipsnames]{xcolor}         %
\usepackage[colorlinks=true,
            linkcolor=Blue,
            urlcolor=Blue,
            citecolor=Blue,
            anchorcolor=Blue]{hyperref}       %

\usepackage{longtable}
\usepackage{multirow}
\usepackage{enumitem}

\usepackage[utf8]{inputenc} 
\usepackage[T1]{fontenc}    
\usepackage{hyperref}       
\usepackage{url}            
\usepackage{booktabs}       
\usepackage{amsfonts}       
\usepackage{nicefrac}       
\usepackage{microtype}      
\usepackage{xcolor}         

\usepackage{tabularx}
\usepackage{array,multirow,graphicx}
\usepackage{multirow}
\usepackage{diagbox}
\usepackage{booktabs}
\usepackage{xcolor}
\definecolor{mygreen}{HTML}{2CB600}
\usepackage{subfig}
\usepackage{xspace}
\usepackage{appendix}
\usepackage{dirtytalk}
\usepackage{epigraph}
\usepackage{makecell}
\usepackage{fancyhdr}
\usepackage{wrapfig}
\usepackage{amsmath}
\usepackage{arydshln}

\usepackage{colortbl}

\definecolor{darkblue}{rgb}{0.0, 0.0, 0.55}
\usepackage{microtype}
\usepackage{makecell}
\usepackage{hyperref}
\usepackage{soul}

\usepackage{algorithm}
\usepackage{algorithmic}

\title{Cracking the Code of Juxtaposition: Can AI Models Understand the Humorous Contradictions}

\author{Zhe Hu$^{1}$\thanks{\quad These authors contribute equally.},  Tuo Liang$^{2}$\footnotemark[1],  Jing Li$^{1}$, Yiren Lu$^{2}$, Yunlai Zhou$^{2}$, Yiran Qiao$^{2}$, Jing Ma$^{2}$, Yu Yin$^{2}$
\\[3pt]
  $^{1}$Department of Computing, The Hong Kong Polytechnic University \\
  $^{2}$Department of Computer and Data
Sciences, Case Western Reserve University
  \\
  $^{1}${\tt zhe-derek.hu@connect.polyu.hk, jing-amelia.li@polyu.edu.hk} \\
  $^{2}${\tt \{txl859,yxl3538,yxz3057,yxq350,jxm1384,yu.yin\}@case.edu}
  \\[3pt]
 {~\texttt{\url{https://vulab-ai.com/projects/yesbut/}}}
  }

\begin{document}

\maketitle

\begin{abstract}
Recent advancements in large multimodal language models have demonstrated remarkable proficiency across a wide range of tasks. 
Yet, these models still struggle with understanding the nuances of human humor through juxtaposition, particularly when it involves nonlinear narratives that underpin many jokes and humor cues.  This paper investigates this challenge by focusing on comics with contradictory narratives, where each comic consists of two panels that create a humorous contradiction. 
We introduce the \textsc{YesBut} benchmark, which comprises tasks of varying difficulty aimed at assessing AI's capabilities in recognizing and interpreting these comics, ranging from literal content comprehension to deep narrative reasoning. 
Through extensive experimentation and analysis of recent commercial or open-sourced large (vision) language models, we assess their capability to comprehend the complex interplay of the narrative humor inherent in these comics. Our results show that even state-of-the-art models still lag behind human performance on this task. Our findings offer insights into the current limitations and potential improvements for AI in understanding human creative expressions.
\end{abstract}

\section{Introduction}
\label{sec:intro}

\vspace{-5mm}
\begin{minipage}[t]{.8\linewidth}
\renewcommand{\epigraphflush}{flushleft}
\setlength{\epigraphwidth}{.9\linewidth}
\epigraph{\itshape ``The world is indeed comic, but the joke is on mankind.''}{---H. P. Lovecraft}
\end{minipage}
\vspace{-4mm}

Comics are a unique blend of visual art and narrative that encapsulate a wide range of human experiences and emotions. 
Understanding comics often requires significant social reasoning skills and cultural knowledge, as they heavily rely on context, cultural references, and visual metaphors. Furthermore, comics frequently employ nonlinear narratives~\cite{pressman2014digital,manning1998understanding}, demanding rigorous reasoning to grasp underlying ideas.
Recent large (vision) language models have achieved impressive performance on various tasks~\cite{liu2023improved,yin2023survey,wu2023multimodal}, yet their ability to comprehend these complex human expressions remains insufficiently explored~\cite{hu2023language,hessel-etal-2023-androids,rayhan2023artificial}.

Examining AI models' ability to understand comics is essential for advancing their social and semantic comprehension. As a significant part of human creative expression, comic offers valuable insights into human emotions and cultural contexts~\cite{duncan2009power}. This understanding is crucial for developing socially intelligent systems and enhancing AI-related creativity, thereby improving user experience in applications such as recommendation systems and automated content creation tools.

Previous studies have applied vision language models (VLMs) to understand humor and deep semantics~\cite{hessel-etal-2023-androids,yang2024can}. 
However, these studies often focus on single-panel comics and do not investigate the more complex case of nonlinear narratives created through juxtaposition, a fundamental element in comics. 
Juxtaposition involves placing two contrasting elements together to provoke thought or evoke humor~\cite{young2003art,groensteen2013comics}.
This technique requires readers to pause and reassess the meaning, engaging in \textit{nonlinear thinking} to reason about the relationships between panels for overall idea.~\cite{bearne2003rethinking,dittmer2010comic,schechter2011juxtaposition}.

\begin{wrapfigure}{tr}{0.52\textwidth}
    \vspace{-3mm}
    \centering
    \includegraphics[scale=0.7]{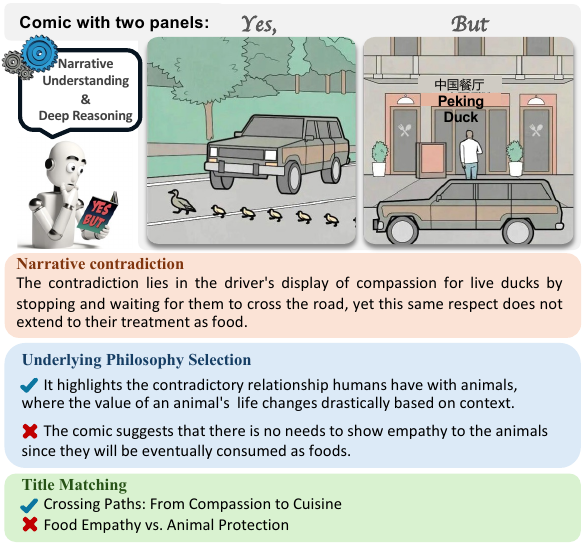}
    \caption{We introduce \textsc{YesBut} dataset for comic understanding of juxtaposed comic panels. Given a two-panel comic with a contradictory narrative, we propose several tasks including narrative understanding, underlying philosophy selection and title matching, tackling different levels of comic understanding. (Comic by Anton Gudim).}
    \label{fig:intro_example}
\end{wrapfigure} 

In this work, we examine VLMs' ability to understand comics, specifically focusing on humor derived from juxtaposition. Our goal is to determine if large models can accurately comprehend the complex and contradictory narratives present in comics. Such contradictions challenge conventional semantic interpretations and demand deeper analysis. For example, Figure~\ref{fig:intro_example} shows a comic with two panels: in the first, a driver stops for ducks to cross the road ("Yes"), and in the second, the driver enters a "Peking Duck" restaurant ("But"), highlighting the contradiction in human-animal relationships through juxtaposition.

Understanding such juxtaposition in comics significant challenges for the models. First, \textbf{it requires deep comprehension of human norms}, recognizing that people often have conflicting feelings, and identifying subtle social cues and contexts tied to cultural backgrounds. Additionally,  \textbf{it demands nonlinear reasoning to grasp the overall narrative}, as the story is conveyed through the interplay of two panel elements, forming the core of the narrative beyond the literal meaning of each single panel. This type of juxtaposition necessitates critical thinking about similarities and differences, requiring in-depth reasoning.   
However, current models lack the ability to process information through nonlinear and deep thinking effectively, as the autoregressive paradigm of LMs limits their bidirectional reasoning capabilities
~\cite{kuttner2021comics,tong2023eliminating,bubeck2023sparks}.
By emphasizing these contradictions, we aim to push AI models to develop more sophisticated semantic understanding, enriching their interpretative capabilities.

To this end, we collected and annotated a new benchmark, \textsc{YesBut}, for understanding comics with juxtaposition, focusing on \textit{contradictory} narratives. Each comic is annotated with a literal description, a contradiction illustration, the underlying philosophy it reveals or satirizes, and a title that summarizes the overall narrative, as shown in Figure~\ref{fig:intro_example}. We then propose four tasks: (1) {literal description writing}, to produce a surface description of the comic narrative; (2) {contradiction generation}, where the model illustrates the narrative contradiction; (3) {underlying philosophy selection}, which targets at selecting the correct philosophy the comic reflects; and (4) {title matching}, where the model matches the comic with a proper title.
These tasks jointly cover different levels of comic understanding, from literal content comprehension to more in-depth narrative reasoning, providing a thorough evaluation of comic understanding capabilities.

We conducted comprehensive experiments on the \textsc{YesBut} dataset, evaluating both commercial and open-sourced large (visual) language models. Both automatic and human evaluation results indicate that commercial VLMs outperform their open-sourced counterparts on most tasks. However, even the highest scores are far from perfect (e.g., 84.1\% accuracy for underlying philosophy selection and 63.3\% for title matching), underscoring the need for further advancements in this area. Additionally, our analysis reveals that augmenting models with oracle comic descriptions can significantly enhance performance, highlighting the considerable gap in current models' understanding of comic narratives. 
We release our annotations, code, and model results, aiming to provide valuable insights for future AI research on understanding human creative expression.

\section{Related Work}
\label{related}
\noindent
\textbf{Large Models and Evaluations.}
Recent large (vision) language models have demonstrated remarkable performance in following human instructions and performing various downstream tasks through zero-shot prompting~\cite{ouyang2022training,llama3modelcard,minaee2024large,yin2023survey}. Various benchmarks have been proposed to evaluate their performance, encompassing both language-only tasks~\cite{zheng2024judging,dubois2024alpacafarm,wang2023pandalm,huang2024c} and vision-language tasks~\cite{ying2024mmt,bitton2023visit,bitton2023breaking,li2023seeda,li2023seedb}. These tasks primarily focus on assessing the fundamental capabilities of large models. However, the ability of large models to perform in-depth social reasoning and accurately understand human contexts remains underexplored~\cite{hu2023language}.

\noindent
\textbf{Computational Humor.} 
Humor is a vital component of human communication~\cite{palmer2003taking}. Our research is closely related to the computational understanding of humor. Previous studies have addressed humor recognition~\cite{chen2017predicting,cattle2018recognizing,yang2015humor} and generation~\cite{amin-burghardt-2020-survey}, with recent work expanding to multimodal data, such as visual humor prediction~\cite{chandrasekaran2016we}, humorous cartoon caption identification~\cite{shahaf2015inside,hessel-etal-2023-androids,radev2015humor}, and humor prediction in videos~\cite{kayatani2021laughing,liu2024comment}.
Despite advancements, recent work shows that LLMs such as ChatGPT has not fully solved computational humor yet~\cite{jentzsch2023chatgpt}.
In this work, we design tasks to evaluate large vision-language models on their ability to understand humor through comic juxtapositions with contradictory narratives, requiring deep narrative comprehension. 
Through this study, we aim to provide insights into the capabilities of AI in processing and appreciating humor.

\noindent
\textbf{Interpretation of Human Creative Expressions.} 
Visual artwork, encompassing mediums such as drawings, paintings, and sculptures, has been a profound aspect of human culture and cognition. These creative expressions are not merely decorative; they are deeply entwined with the ways humans perceive, interpret, and communicate their experiences and emotions~\cite{de2016brain}.
Understanding these human creative expressions necessities valuable insights of human emotions, societal values, and cultural contexts, which
is crucial for developing socially intelligent systems and enhancing AI-related creativity~\cite{koivisto2023best}. Previous research has explored AI interpretation of visual human creative expressions in tasks such as meme~\cite{hwang-shwartz-2023-memecap} and cartoon~\cite{radev-etal-2016-humor} understanding. Similar to our work, studies like~\cite{hessel-etal-2023-androids} and~\cite{yang2024can} apply AI models to comprehend comics. However, these studies primarily focus on single-panel comics, emphasizing humor and deep semantics. In contrast, our work aims to investigate the significant feature of juxtaposition for understanding contradictory narratives.

\noindent
\textbf{Visual Reasoning.}
Our task is also related to the visual reasoning ability, where the model requires in-depth reasoning to comprehend contradictions between two comic panels. Previous research has examined visual reasoning capabilities of large models in tasks involving commonsense reasoning~\cite{wang2023gemini,bitton2023breaking,zellers2019recognition}, visual question answering~\cite{hudson2019gqa}, visio-linguistic compositionality~\cite{thrush2022winoground}, and science question answering~\cite{lu2022learn}. 
Unlike these studies, our task involves nonlinear reasoning, which necessitates AI to navigate multi-dimensional and complex information layers, often without explicit directives. While linear reasoning present their challenges, they usually exhibit clearer rules and structures, making them more accessible for AI to process with existing algorithms and models. Consequently, nonlinear reasoning represents a more intricate task, demanding higher natural language processing and cognitive modeling capabilities from AI systems.

\section{The \textsc{YesBut} Dataset}
\label{dataset}

Our benchmark consists of \textsc{YesBut} comics featuring contradictory narratives. Specifically, each sample includes: (1) a two-panel comic that forms a narrative with inherent contradictions; (2) a literal description of the comic narratives; (3) an explanation that illustrates the contradiction within the narrative; (4) the deep philosophy or underlying message the comic aims to convey; and (5) a title of the comic. Based on these components, we construct various tasks for comic understanding.

\subsection{Image Collection}
Our dataset consists of captionless comics, primarily from Anton Gudim's "YES, BUT" series~\cite{gudimtwitter}, each featuring two panels depicting contradictory everyday scenarios. We scraped the images from social media~\footnote{\url{https://twitter.com} and \url{https://www.pinterest.com/}} and conducted preprocessing, including deduplication, filtering out comics with more than two panels, and removing any inappropriate or offensive content. This process resulted in a final dataset of 348 comics.

\begin{figure*}[t]
    \centering
    \includegraphics[scale=0.56]{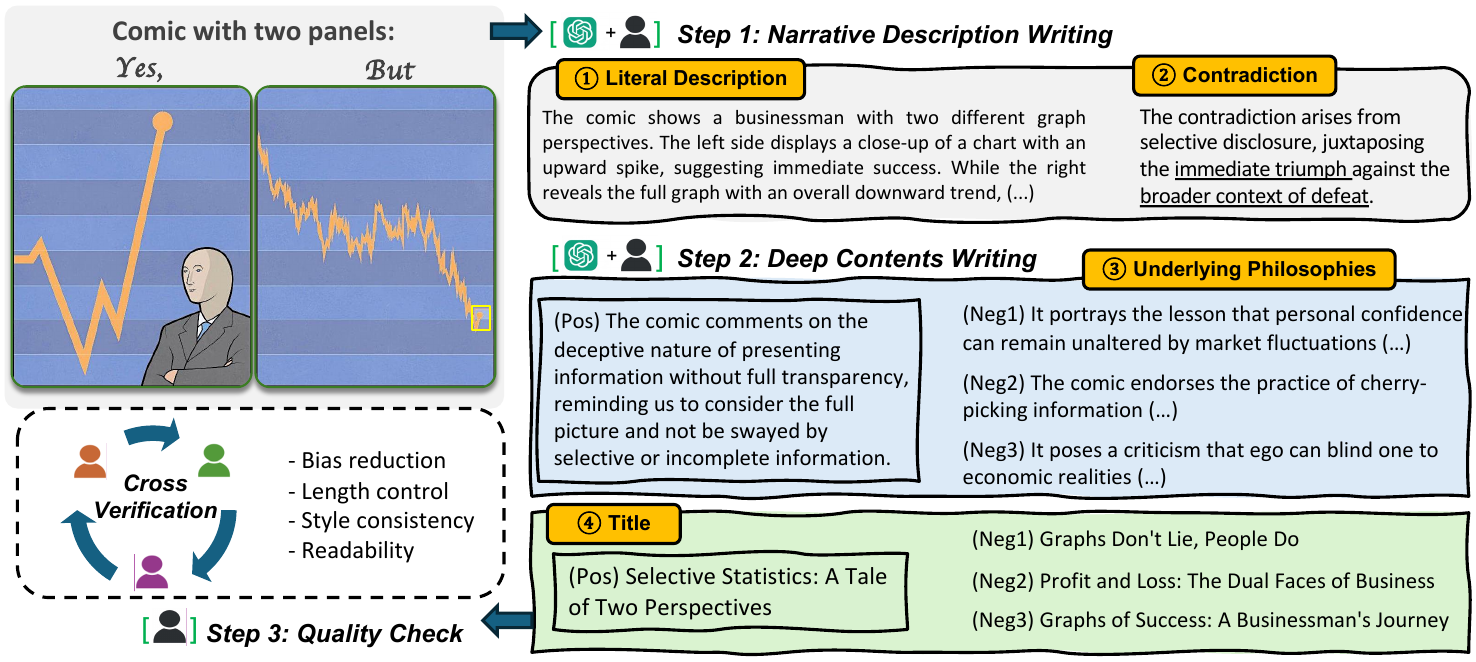}
    \captionof{figure}{Overview of the data construction pipeline. Pos represents the positive options, and Neg stands for the negative options.
    }
    \label{fig:annotation_pipeline}
\end{figure*}

\subsection{Data Annotation}
For each comic, we annotate the corresponding literal description, contradiction explanation, underlying philosophy and comic title~\cite{hessel-etal-2023-androids,yang2024can}. We primarily rely on human annotators to obtain gold-standard annotations. Eight human judges participated in the annotation process, all of whom are proficient English speakers based in English-speaking countries and have at least a Bachelor's degree. Our annotation process included two stages: the progressive human-AI collaborative annotation stage and the quality check and cross-verification stage. The pipeline is illustrated in Figure~\ref{fig:annotation_pipeline}.

\noindent\textbf{Progressive Human-AI Collaborative Annotation.} 
In this stage, we randomly assigned comic samples to each annotator, instructing them to first exclude any comics that may contain offensive, hateful, or sexual material before beginning the annotation process. To reduce human effort and costs associated with data annotation from scratch, we designed a human-AI collaboration pipeline utilizing GPT-4~\cite{achiam2023gpt} for data annotation and component writing. 

The pipeline operates through dialogue interactions with the GPT-4 model. Given a comic image, we first prompt GPT-4 to generate narrative descriptions, illustrating the comic's narrative and explaining the contradictory logic between the two panels. Human annotators then modify and annotate the contents to obtain a literal description and contradiction explanation. 

After obtaining the gold-standard description, both the comic and the description are used as input to prompt GPT-4 deep content writing, including the underlying philosophy and an eye-catching comic title. The underlying philosophy aims to foster a deep understanding of the comic and reveal the phenomenon it satirizes or the lesson it conveys; and the title is a more abstractive expression that reflects the overall narrative. Both components will be further checked by human annotators. Additionally, for the underlying philosophy and title understanding tasks, GPT-4 generates hard negative counterparts and distractions to design multiple-choice questions for our experiments. The prompts we used are provided in the Appendix~\ref{sec:annotation_details_appendix}.

Our human-AI collaborative annotation pipeline is effective as it leverages a progressive prompting strategy, annotating each component from easy to difficult. Understanding the underlying philosophy of the comic, for example, requires first understanding the literal narratives and contradictions. This approach reduces annotation costs and improves overall efficiency.

\noindent\textbf{Quality Check with Cross Verification.} 
To ensure the quality and accuracy of the components and reduce objective bias from different human annotators, we introduced a cross verification stage. 
In this process, one annotator is assigned as an inspector for each comic. The inspector checks the annotated results to ensure all components are correct, unbiased, and appropriate. If any content is found to be of low quality or ambiguous, a third annotator is brought in as a judge to determine the final version. We exclude the comics with ambiguous or controversial narratives. This process ensures the quality of the annotated components for benchmark construction.
\begin{wraptable}{r}{0.38\textwidth}
\centering
\small
    \begin{tabular}{@{}lcc@{}}
        \toprule 
         Components & \makecell{\#Num} & \makecell{Avg. Len.}   \\
         \midrule 
          \rowcolor{lightgray!30}
        Image & 348 & - \\
        Literal Description & 348 & 80 \\
        Contradiction & 348 & 31 \\
        Philosophy  & 1,392 & 24 \\
        Title & 1,392 & 6 \\
         \bottomrule
    \end{tabular}
    \caption{
    Data Statistics. Avg. Len. is the average number of words.
    }
    \label{tab:data_statistics}
\end{wraptable} 
After the cross-verification process, one of the authors reviews each sample to verify the validity and appropriateness of the components. 
Finally, we obtain a dataset of 348 comics, each accompanied by high-quality components. The statistics of the components are shown in Table~\ref{tab:data_statistics}.

\noindent\textbf{Mitigating Annotation Bias.}
Our benchmark focuses on \textit{common interpretation} of humor. However, the subjectivity of this task may introduce bias. To mitigate this issue, we have taken several steps in our annotation process: 
(1) Our annotators come from different genders and diverse cultural backgrounds, providing a range of perspectives;
(2) Multiple quality checks and verifications are incorporated to ensure consensus among different annotators, with controversial or potentially biased comics being filtered out;
(3) Annotations are further validated by cross-referencing social media comments for each comic to ensure alignment with widely accepted interpretations;
(4) Recognizing that tasks such as generating titles and philosophical contents are inherently open-ended and involve subjective data annotation, we frame them as selection tasks, and ensure that the correct option is clearly and objectively superior than the negative options to mitigate subjectivity.

\subsection{Task Design: Do Large Models Understand Humor in Juxtaposition?}
We aim to evaluate the capabilities of recent large (visual) language models in understanding humor through contradictions. This is challenging because it requires both social reasoning about human events and nonlinear logical reasoning about the narratives, going beyond the literal understanding of the comic. We design a series of tasks that require different levels of narrative understanding and reasoning abilities to evaluate the models' performance in reading comics.

\noindent\textbf{1. Literal Description Writing.} 
The first task is to generate the literal description of the comic narrative. We formulated this task as a text generation task: given an input comic, the model is required to generate a short description illustrating the narrative from the two panels of the comic. This task is different from the traditional image captioning, which requires the model to illustrate the comic narrative instead of solely focusing on image description.

\noindent\textbf{2. Contradiction Generation} evaluates whether the model can understand the contradiction within the narrative juxtaposition. Similarly, it is formulated as a text generation task.

\noindent\textbf{3. Underlying Philosophy Selection.}
Understanding comics requires grasping not only the surface meaning of the images but also the underlying ideas the authors aim to convey. This task evaluates the model's ability to recognize the comic's underlying philosophy. It is formulated as a multiple-choice question answering (MCQ) task: given an input comic and four candidates of its underlying philosophy, the model must predict the correct option. The negative choices are crafted by annotators to be relevant to the comic, requiring reasoning to make the correct prediction.

\noindent\textbf{4. Title Matching}
evaluates whether the models can identify the corresponding title, which is challenging because the title acts as an abstraction of the narrative and requires a proper understanding of the comic's content. Similar to the underlying philosophy task, it is formulated as a multiple-choice question answering task, where the most is asked to select the correct title from four options.

\section{Experiments}
\label{exp_setting}

\begin{table*}[t]
\fontsize{9}{12}\selectfont
\setlength{\tabcolsep}{1.6mm}
\centering
\begin{tabular}{@{}ll| ccc ccc c c@{}} \toprule
\multicolumn{2}{l}{\multirow{2}{*}{}}   
& \multicolumn{3}{c}{\textbf{Literal Description}}
& \multicolumn{3}{c}{\textbf{Contradiction}} & \textbf{Philosophy} & \textbf{Title}    \\ \cmidrule(l){3-5} \cmidrule(l){6-8} 
\cmidrule(l){9-9} \cmidrule(l){10-10} 
\textbf{Setting} & \textbf{Model} & 
$\text{BERT}$  & R-2 & GPT  &
$\text{BERT}$  & R-2 & GPT   & Accuracy & Accuracy   \\
\midrule
\multirow{10}{*}{\makecell{VLMs}}  
& {GPT-4}  &  \textbf{88.32} & \textbf{87.46} & \textbf{3.76} & \underline{87.64}  & \textbf{83.21}  & \textbf{4.03} &  \underline{82.76}  & \underline{60.25} \\
& {Claude-3} &  \underline{87.68} & \underline{80.30} & \underline{3.28} & 86.93 & 80.63   & \underline{3.79}  & \textbf{84.10} & 56.42  \\
& {LLaVA-1.6-34B} & 86.45  & 67.67 & 2.86 & 86.04 & 75.95   &  3.51 & 78.83 & \textbf{63.31} \\
& {LLaVA-1.6-13B} & 81.34  & 75.95 & 2.96 & 86.48   & \underline{80.96} & 3.36  & 69.16 &  55.08 \\
& {LLaVA-1.5-13B} & 78.77  & 58.21 & 2.51 & 86.48   & 67.67 & 3.36  & 69.73 & 48.75  \\
& {InstructBlip-13B} &  85.20 & 35.28 & 2.69  & 85.54   & 51.15 & 2.54  & 30.75 & 22.70 \\
& {CogVLM} & 80.80  & 55.51 &  2.65 & {87.07}   & 69.96 & 3.76  & 61.30 & 49.52 \\
& {Qwen-VL-Chat} & 79.03  & 51.58 & 2.76 & 86.41 & 59.77 & 3.25  & 59.10 & 42.05 \\
& {mPlug-Owl2} & 78.26  & 47.38 & 2.57 & 86.20 & 48.05 & 2.59  & 62.17 & 43.10  \\
& {LLaVA-1.6-7B} &  80.71 & 70.36 & 2.79 & 86.58 & 75.36 & 3.24  & 47.41 & 37.07  \\
& {InstructBlip-7B} &  76.02 & 38.02 & 2.60 & 86.32 & 66.29 & 2.85  & 25.86 & 26.44 \\
\midrule
\multirow{3}{*}{\makecell{LLMs}}  
& {ChatGPT}  & -  &-  & - &  \textbf{87.78}  & 67.42 & 3.54 & 75.86 & 49.52  \\
& {Llama-3-8B-Instruct}  & -  &-  & -  & 87.41   & 70.52 & 3.59  & 72.13 & 49.71 \\
& {Mistral-7B-Instruct} &-   &-  & -   &  87.01   & 67.70 &  3.64 & 66.00 & 45.98\\
\bottomrule
\end{tabular}
\caption{Main results. For literal description and contradiction, we report BERT score (recall), BLEURT (BLT), and GPT evaluation score. For philosophy and title, we report accuracy (\%).
Best scores are \textbf{bold} and the second best ones are marked with \underline{underline}.}
\label{tab:main_results}
\end{table*}
\subsection{Models and Settings}

We evaluate the models' performance in a zero-shot manner using both recent VLMs and LLMs. For VLMs, the comic image and questions are provided as inputs for output prediction. We include both commercial models such as GPT-4~\cite{achiam2023gpt} and Claude-3\cite{anthropic2024claude}, as well as open-sourced models including LLaVa~\cite{liu2023improved,liu2024llavanext}, CogVLM~\cite{wang2023cogvlm}, Qwen-VL~\cite{Qwen-VL}, mPLUG-Owl2~\cite{ye2023mplugowl2}, and InstructBLIP~\cite{dai2024instructblip}.

For LLMs, since they cannot directly process images, we use the LLaVa-1.6 13B model generated literal descriptions as inputs due to its strong performance.  We include ChatGPT, the Llama3 instruction model~\cite{llama3modelcard}, and the Mistral 7B instruction model~\cite{jiang2023mistral}. More details of the models are included in Appendix~\ref{sec:exp_detail_appendix}.

\noindent\textbf{Implementation Details.}
For GPT-4 and ChatGPT, we set the temperature as 1. For other models, we use the default parameter settings during inference.
To reduce variance across different task prompts, we create three distinct prompts for each task and report the average scores from three runs with each prompt. The specific prompts and additional details are in Appendix~\ref{sec:exp_detail_appendix}.

\subsection{Evaluation Metrics}

For the philosophy and title understanding tasks, which are formulated as multiple-choice question answering, we use accuracy as the evaluation metric. For generation tasks including literal description and contradiction, we apply reference-based evaluation metrics commonly used in text generation studies~\cite{celikyilmaz2020evaluation}, and report ROUGE-2 (recall)~\cite{lin-2004-rouge} and BERT Score (recall)~\cite{bert-score}.~\footnote{We report recall scores considering the open-ended nature of the outputs, as there can be multiple valid expressions. Our focus is on evaluating whether the key points are covered by the model outputs, ensuring a more precise assessment of content coverage.} Recent work shows GPT-based evaluation aligns well with human judgements~\cite{chan-etal-2023-clair,liu-etal-2023-g,hu2023americano}, and we also apply ChatGPT for evaluation~\footnote{
We utilize different GPT variants for specific purposes: \textit{gpt4-turbo} for data annotation, \textit{gpt-4-vision-preview} for experiments, and \textit{gpt-3.5-turbo-0125} for GPT-based evaluation in text generation tasks. This helps reduce potential evaluation bias toward GPT-4's own generation. Further details are in Section~\ref{sec:exp_detail_appendix}.}. The prompts for GPT-based evaluation are provided in Appendix~\ref{sec:exp_detail_appendix}.

Due to the limitation of the automatic evaluations for text generation, we also include human evaluation to assess the quality of the outputs for the literal description and contradiction generation tasks. We hire three human judges to rate each aspect on a scale of 1 (worst) to 5 (best). For literal description, following \cite{hwang-shwartz-2023-memecap}, we evaluate: (1) \textbf{Correctness}: Does the model output correctly convey the narrative of the comic? (2) \textbf{Completeness}: Does the model output cover all the important elements of the comic narrative? (3) \textbf{Faithfulness}: Can all contents from the model output be supported by the comic image (i.e., there are no hallucinations)? For contradiction generation, we evaluate \textbf{Correctness} and \textbf{Faithfulness}. More details are provided in Appendix~\ref{sec:human_eval_details}.

\section{Main Results}
\label{results}

\begin{figure*}[t]
    \centering
    \includegraphics[scale=0.22]{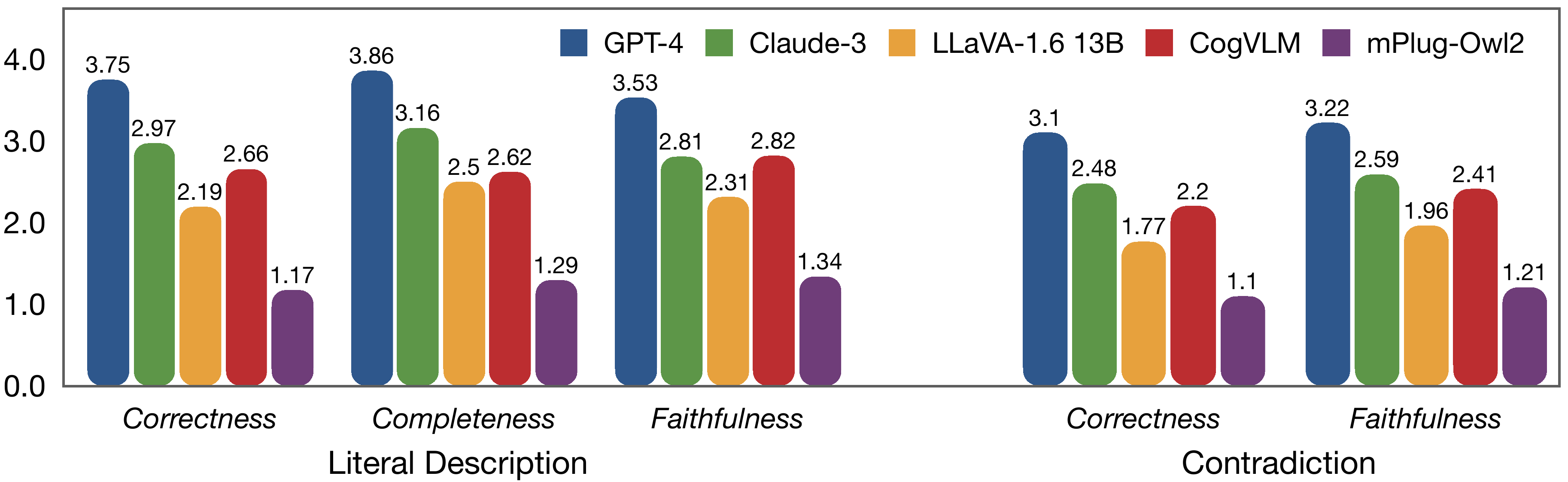}
    \captionof{figure}{Human Evaluation on literal description and contradiction generation.
    }
    \label{fig:human_eval}
\end{figure*}

The main experimental results are shown in Table~\ref{tab:main_results}. For VLMs, the original image is directly used as input, while for LLMs, the generated comic description is used as input.
\subsection{Narrative Understanding Tasks}
\textbf{Literal Description:} We evaluate the results of VLMs only for this task. We observe that the two commercial models generally outperform the smaller open-sourced models. Among these models, GPT-4 achieves the highest scores.  For the open-sourced models, the larger model variants (13B) consistently achieve better scores than their 7B counterparts, indicating that larger models have a superior ability to understand the image and produce higher-quality literal descriptions.

\textbf{Contradiction Generation:} A similar trend is observed where GPT-4 and Claude-3 achieve better results than other VLM models. Notably, LLaVA-1.6 variants outperform their counterparts in generating contradiction descriptions. This is likely due to their improved reasoning ability and world knowledge~\cite{liu2024llavanext}, which are essential for understanding comic narratives and accurately capturing the relationship between the two panels. For LLMs, unlike VLMs, the Llama-3 and Mistral models achieve results comparable to ChatGPT. Another interesting observation is that Llama-3 and Mistral obtain similar or better results for contradiction generation compared to open-sourced VLMs, despite not having access to the original comic images.


\subsection{Deep Reasoning Tasks}

The Underlying Philosophy Selection and Title Matching tasks require in-depth reasoning based on the comic narratives. As seen in Table~\ref{tab:main_results}, for philosophy selection, Claude-3 achieves the best accuracy with 84.10\%, while for title matching, the LLaVA-1.6 34B variant ranks the highest with 63.31\% accuracy. One key observation is that larger models usually perform better in-depth understanding of the comics, aligning with the findings that larger models typically exhibit superior reasoning abilities~\cite{qiao-etal-2023-reasoning,wei2022emergent}.

Additionally, LLMs achieve performance comparable to open-sourced VL models. This can be attributed to the strong reasoning abilities of models like Llama-3 and Mistral~\cite{jiang2023mistral,llama3modelcard}, which are crucial for understanding narratives and performing nonlinear reasoning to grasp deep semantics. Further analysis on the influence of descriptions for LLMs is provided in Section~\ref{sec:llm_caption_influence}.

Another observation is that model performance on title matching is consistently lower than on underlying philosophy selection. Titles are shorter and more abstract versions of the narrative and do not explicitly convey the underlying idea of the comic. Therefore, distinguishing the correct title from distractions requires a deeper rigorous understanding and reasoning abilities, making it more challenging for models. Notably, the human evaluation results show a similar trend of our proposed GPT-based evaluation, demonstraining its effectiveness.

\subsection{Human Evaluations}

We conduct human evaluations on 30 randomly selected samples to assess the output quality of literal descriptions and contradiction generation, as shown in Figure~\ref{fig:human_eval}. Similar trends are observed in both human and automatic evaluations: commercial models generally outperform open-source models in producing both literal descriptions and contradictions, with GPT-4 achieving the highest scores in both tasks. Additionally, the scores for literal descriptions are consistently higher than those for contradictions across all models, suggesting that understanding narrative contradictions is more challenging than generating literal descriptions, which requires in-depth reasoning to compare the various aspects of both panels.

A comparison of the scores for literal description and contradiction reveals a strong correlation between the two tasks: models that perform well on literal descriptions also tend to achieve good results on contradictions. This indicates that understanding comic juxtaposition requires a diverse set of skills, including image understanding, narrative comprehension, and reasoning abilities.

\section{Analysis and Discussion}
\label{analysis}

\subsection{How Does Literal Understanding of the Comic Influence Deep Reasoning?}
\label{sec:llm_caption_influence}

\begin{figure}[t]
\begin{minipage}[b]{.48\textwidth}
\centering
\includegraphics[width=0.8\textwidth]{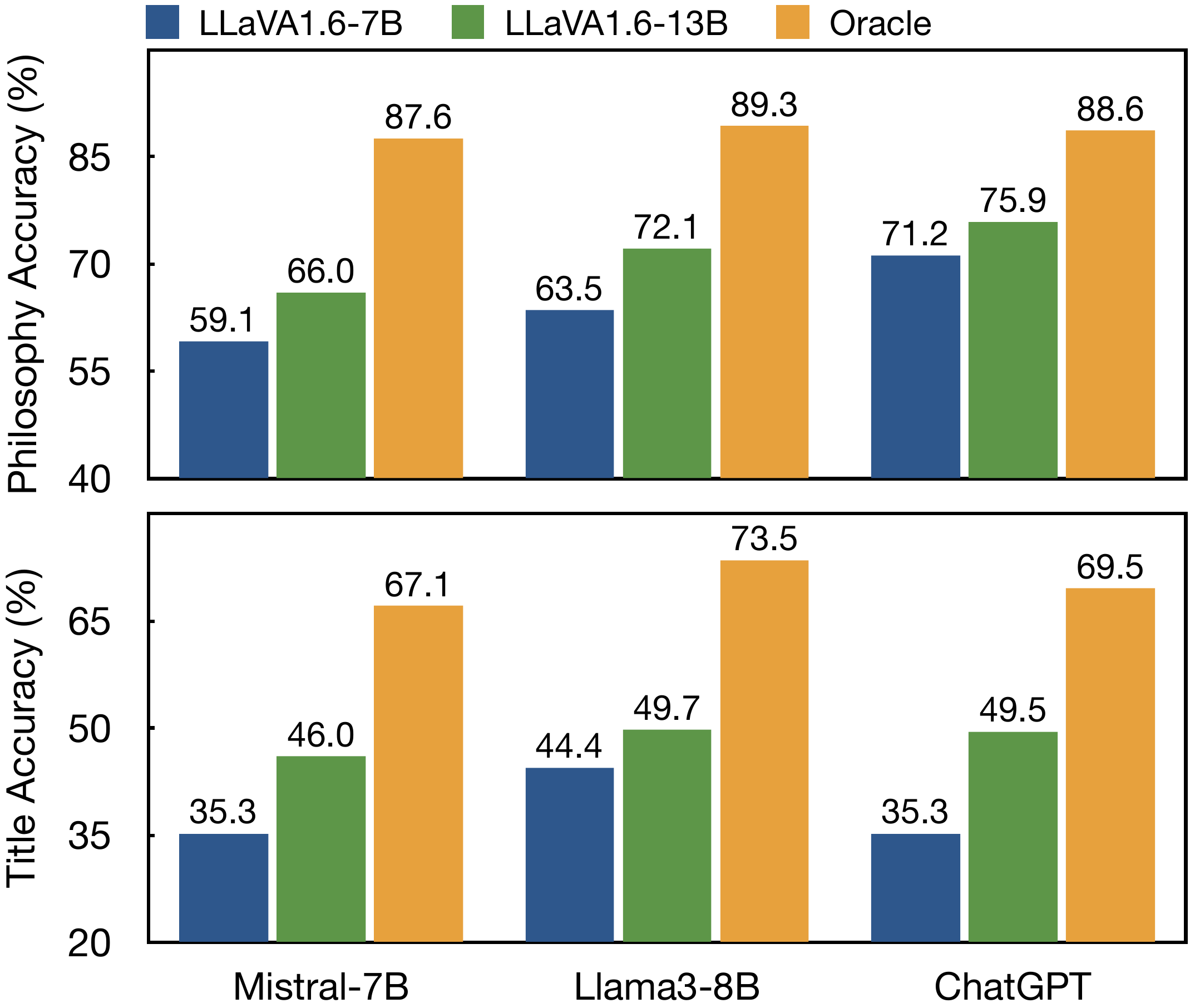}
\caption{LLMs using different image description as input.}
\label{fig:llm_caption}
\end{minipage}
\hfill
\begin{minipage}[b]{.48\textwidth}
\centering
\includegraphics[width=0.8\textwidth]{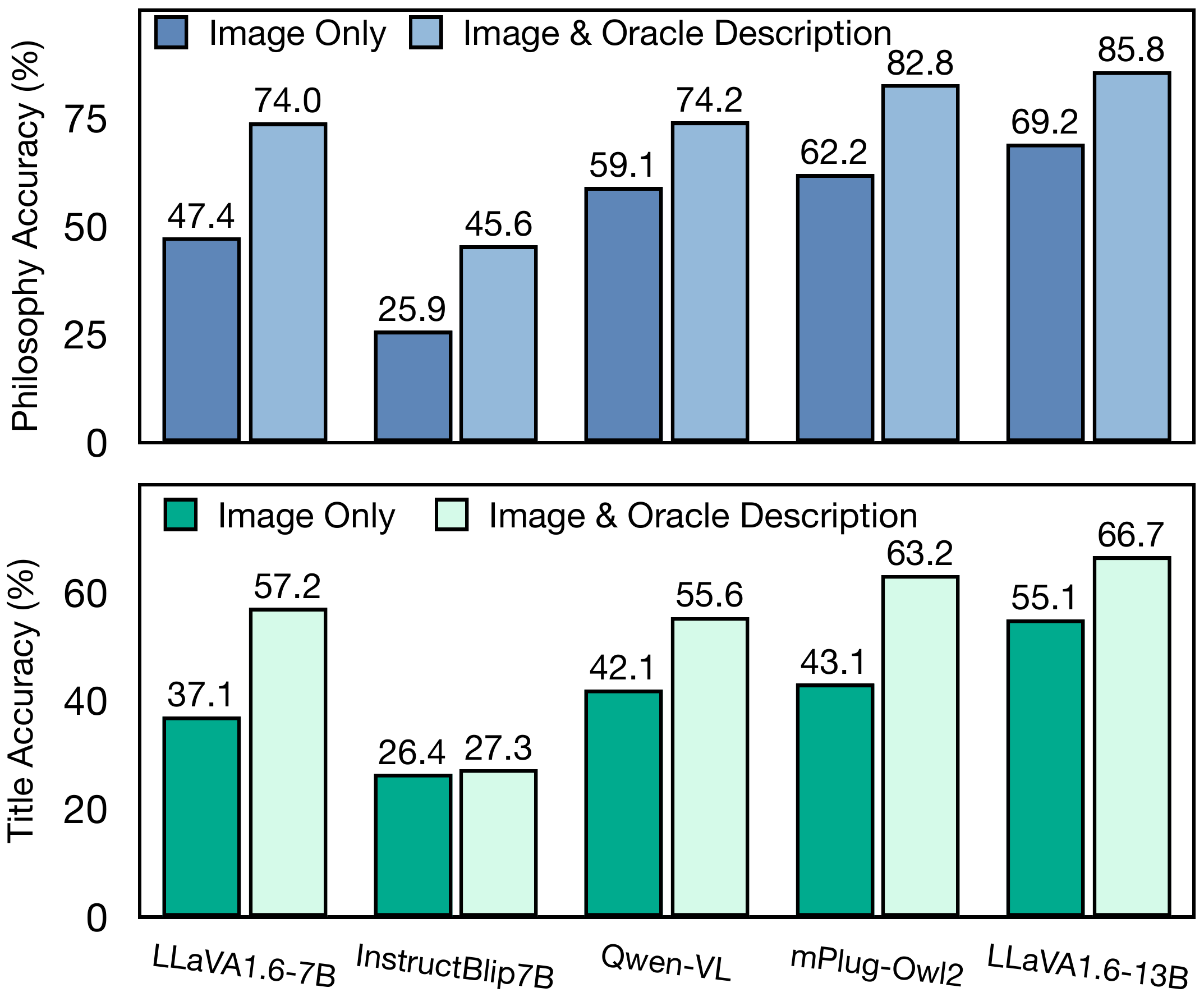}
\caption{VLMs with image only input and image + oracle description as inputs.}
\label{fig:vlm_caption}
\end{minipage}
\end{figure}

We investigate whether the quality of surface-level literal descriptions influences subsequent deep reasoning tasks. For LLMs, we provide different literal descriptions generated by LLaVA-1.6 7B and 13B variants, as well as oracle descriptions written by humans, as model inputs. The results are shown in Figure~\ref{fig:llm_caption}. As the quality of literal descriptions improves, the prediction accuracy for both underlying philosophy and title selection also improves. This demonstrates a strong correlation between deep reasoning and literal narrative understanding. However, a significant performance gap remains compared to when oracle descriptions are used.

We further examine the performance of VLMs by providing them with additional oracle descriptions. The results are shown in Figure~\ref{fig:vlm_caption}. Compared to using only the comic image as input, augmenting with human-written literal descriptions significantly improves the deep reasoning results for all VLMs. This confirms that correctly reasoning about the underlying semantics of a comic requires first accurately understanding its surface narrative. However, the performance gap indicates that current VLMs still lag in narrative understanding.

Additionally, an interesting observation from Figures~\ref{fig:llm_caption} and \ref{fig:vlm_caption} is that when the oracle literal description is provided as (partial) input, LLMs tend to outperform their VLM counterparts in both philosophy and title selection. For example, LLaVA-1.6-7B employs Mistral-7B as the language model backbone, yet its performance under oracle description is significantly worse than that of Mistral-7B . One possible reason is that incorporating oracle descriptions makes VLM input much longer, thus making prediction more challenging. We provide further discussions in Section\ref{sec:decomposition}.

\subsection{Is Decomposing Literal Description Helpful for Deep Reasoning of VLMs?}
\label{sec:decomposition}

VLMs typically predict results in an end-to-end fashion, requiring the model to perform image captioning, narrative understanding, and deep reasoning all at once. Here, we investigate whether decomposing the task into separate stages of narrative understanding and in-depth reasoning can improve model performance. Specifically, we first prompt the VLM to produce a literal description of a comic; then the VLM predicts results based on both the comic image and the description. The results are shown in Table~\ref{tab:decomposed_vlm}.

\begin{wraptable}{r}{0.38\textwidth}
\centering
\small
    \begin{tabular}{@{}lcc@{}}
        \toprule 
         Models & \makecell{Philosophy} & \makecell{Title}   \\
         \midrule 
          \rowcolor{lightgray!30}
         LLaVA-1.6-13B & 69.16 & 55.08  \\
        \quad\quad$\hookrightarrow$ w/ desp. & 68.68 & 48.76 \smallskip\\  
        Qwen-VL-Chat & 59.10 & 42.05 \\
        \quad\quad$\hookrightarrow$ w/ desp. & 59.58 & 37.55 \smallskip \\  
        mPlug-Owl2  & 62.17 & 43.10 \\
        \quad\quad$\hookrightarrow$ w/ desp. & 60.25 & 37.84 \smallskip\\  
        LLaVA-1.6-7B &47.41  & 37.07 \\
        \quad\quad$\hookrightarrow$ w/ desp. & 53.07 & 34.96 \\  
         \bottomrule
    \end{tabular}
    \caption{
    Decomposition model results augmenting the predicted description.
    }
    \label{tab:decomposed_vlm}
    \vspace{-12pt}
\end{wraptable} 

As observed, decomposing the task and augmenting it with a literal description does not necessarily improve performance. In fact, when descriptions are incorporated, performance across all models declines on the title selection task, which contrasts with previous findings~\cite{yang2024can}. One possible explanation for this drop in performance is that the generated descriptions may contain errors, negatively impacting the model's deep understanding. Another explanation could be the length of the generated descriptions (e.g., the LLaVA-1.6 13B model's descriptions average around 170 words), leading to longer and more complex prompts that make prediction more challenging. We leave a more detailed investigation of this issue for future work.


\begin{figure*}[t]
    \centering
    \includegraphics[scale=0.67]{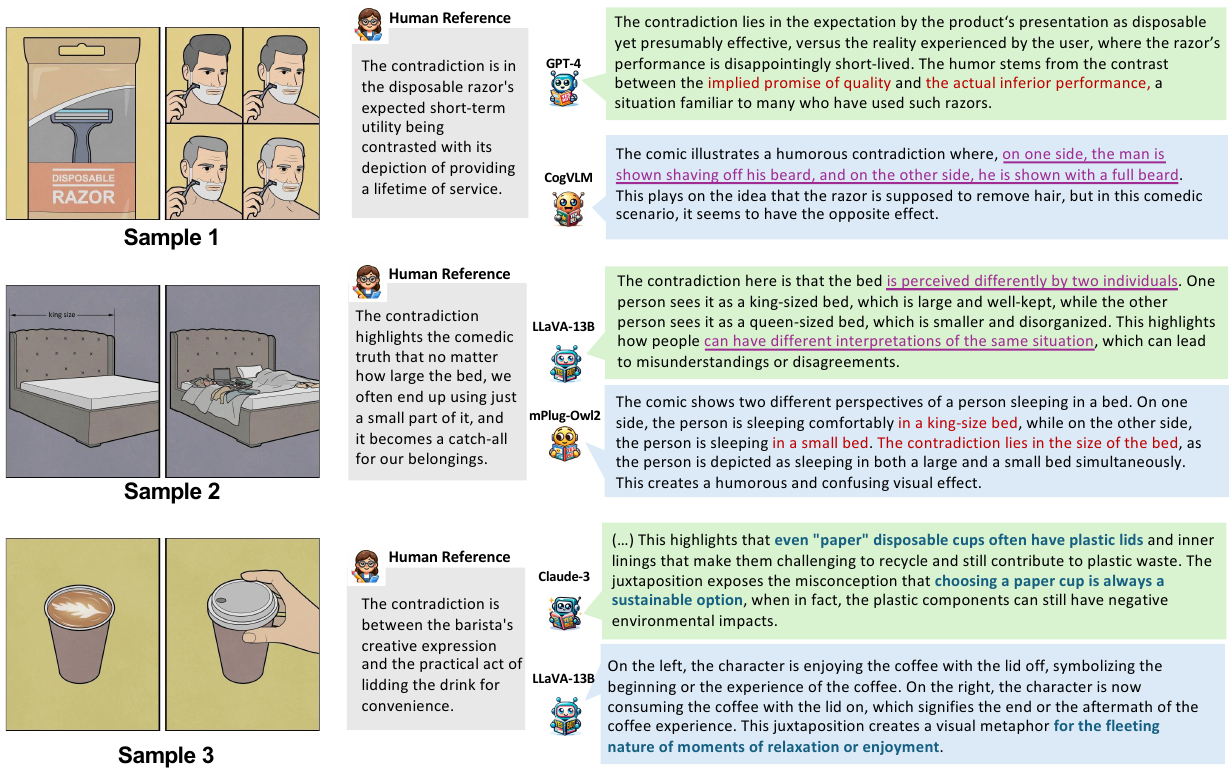}
    \captionof{figure}{Sample outputs of contradiction explanations generated by different vision language models, along with human written references. We highlight different types of errors in model outputs.
    }
    \label{fig:case_contraidction}
\end{figure*}

\subsection{Error Analysis and Future Directions}
We present sample outputs of contradictions generated by vision language models (VLMs) in Figure~\ref{fig:case_contraidction}. VLMs can make various errors in contradiction understanding.

One type of error is \textbf{\textcolor{purple}{\underline{visual misinterpretation}}}, where the model incorrectly interprets the image contents. For example, in sample 1, CogVLM misinterprets the image by recognizing a person "shown with a full beard." Similarly, in sample 2, LLaVA-1.6 13B misunderstands the image contents and generates incorrect content about "two individuals," which is inconsistent with the comic. Such misinterpretations can lead to incorrect understanding of the narrative. These observations align with our previous findings in Section~\ref{sec:llm_caption_influence} and Section~\ref{sec:decomposition}.
\textit{This highlights the need for future research to improve models' visual interpretation capabilities}.

Models also struggle to conduct \textbf{\textcolor{red}{in-depth reasoning of the relationship}} between two panels by recognizing their differences and similarities. In sample 1, while the comic implies a comparison between the expected disposable razor and its actual longevity, GPT-4 incorrectly explains the contradiction as being about the razor's quality. A similar error occurs with mPlug-Owl2 in sample 2, where it incorrectly thinks the bed sizes are different in the two panels, leading to a wrong illustration focusing on the bed size. \textit{Future work might incorporate recent advanced reasoning approaches (e.g., multi-agent debate~\cite{du2023improving}, test-time compute scaling~\cite{snell2024scaling}) to further improve model performance.}

Another common error is \textbf{\textcolor{darkblue}{hallucination and incorrect association}}. This is evident in sample 3. The original comic contrasts latte art before and after lidding the drink, but Claude-3 incorrectly associates the narrative with environmental protection, focusing on the plastic lid. Meanwhile, LLaVA-1.6 13B model suffers from hallucinations by interpreting the narrative as being about relaxation and enjoyment, which is unsupported by the original comic. \textit{This suggests the need for improving world knowledge and social understanding abilities to enhance model performance on this task.}
More sample outputs are in Appendix~\ref{sec:sample_outputs_appendix}.

\section{Conclusion}
In this work, we present \textsc{YesBut}, the first benchmark dedicated to studying comic understanding through juxtaposition. \textsc{YesBut} encompasses a variety of tasks that address both narrative comprehension and deep reasoning. The results indicate that state-of-the-art vision and language models still struggle with these tasks. We also offer a comprehensive analysis and discussion of errors to evaluate model performance. Current models still struggle to accurately interpret the visual contents and conduct in-depth reasoning of the underlying narratives.
Through this study, we aim to provide insights for future research and advance the capabilities of AI models in understanding human context, ultimately contributing to more effective and culturally aware AI applications.

\section{Limitations}
\label{sec:limitation}
We propose a comprehensive data annotation process to annotate each component. However, due to the subjectivity of comic interpretation, especially regarding the underlying ideas, there might be potential ambiguity. While we acknowledge the relatively small size of images, we rigorously collect comics and annotate each component, ensuring their high-quality and reliability. We plan to expand the dataset with the inclusion of different types of narratives in future work.

Our proposed benchmark focuses predominantly on recognizing and interpreting visual humor via juxtaposition, and may not cover all aspects of visual understanding required for more generalized AI applications.
In the future, we intend to explore more deeply how AI can not only interpret but also creatively engage with content. This includes generating pivotal turning points from one perspective and creating counterpoints to given scenarios, like generating a "YES" image's counterpart.

\section{Ethics Statement}
\label{sec:ethics}
\textbf{Copyright and License.} 
All data samples collected are sourced from publicly available content on social media platforms. We ensure compliance with copyright by utilizing original links to comics without infringement. In addition, we obtained permission from the author artist (e.g., {Anton Gudim}) to conduct our benchmark using these public images. Additionally, we commit to open-sourcing our annotated benchmark, providing corresponding links to each comic image. We diligently review samples, filtering out potentially offensive or harmful content.

\textbf{The Large Vision Language Models}  utilized in our experiments are pretrained using diverse web corpora, which may introduce biases in their outputs. We advise users to conscientiously evaluate the ethical implications of generated outputs when employing them in future research endeavors.

\textbf{Data Annotation.} Eight human judges are engaged in our annotation process. We compensate these judges with an average hourly wage of \$11, ensuring fair remuneration for their contributions.

\section*{Acknowledgements}
\label{sec:acknowledgements}

This work made use of the High Performance Computing Resource in the Core Facility for Advanced Research Computing at Case Western Reserve University, which is supported by NSF award NSF-2117439.
We also thank the support from OpenAI Researcher Access grants \#0000007745.

\bibliography{custom}

\newpage
\appendix


\section{Data Annotation Details}
\label{sec:annotation_details_appendix}

\begin{figure*}[ht]
    \centering
    \includegraphics[scale=0.7]{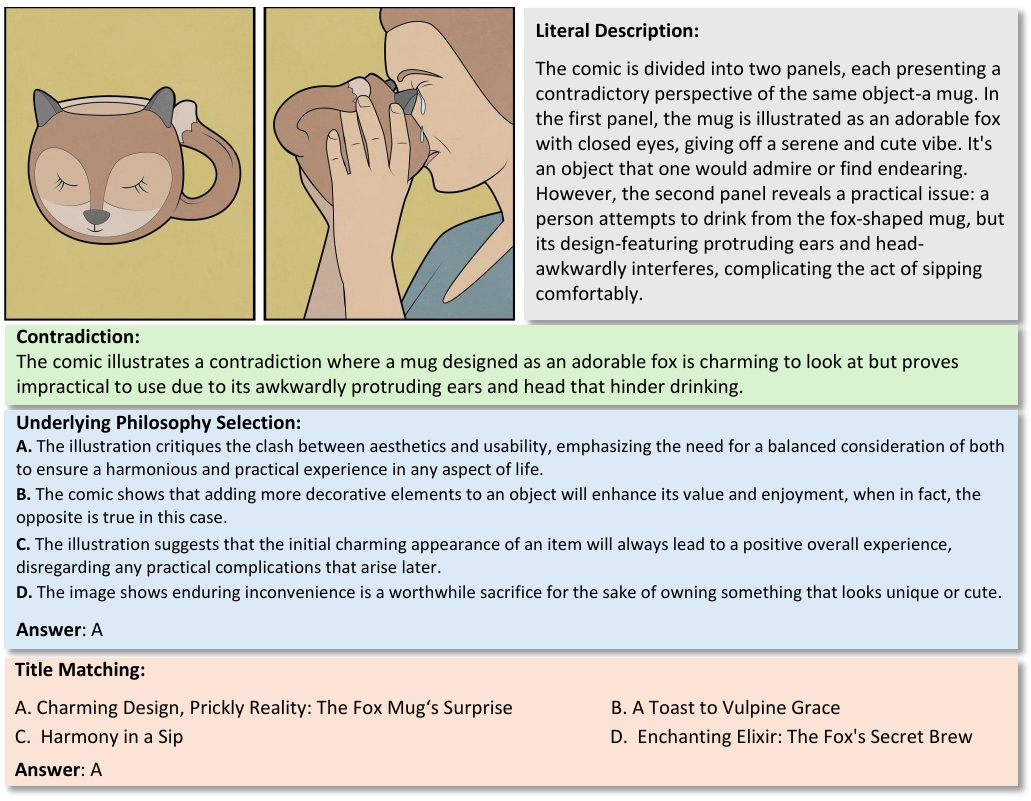}
    \vspace{-2mm}
    \captionof{figure}{Sample comic with all annotated tasks.
    }
    \vspace{-2mm}
    \label{fig:task_samples}
\end{figure*}

Considering the workload of manually writing all components from scratch, we leverage a AI-human collaborative pipeline for annotation. The prompts for generating each component are listed in Table~\ref{tab:annotation_prompts}. After producing each component, human annotators will verify and modify the outputs for the final components. We present a sample comic with all tasks in Figure~\ref{fig:task_samples}.

\begin{table}[ht]
\centering
\scalebox{0.9}{
\fontsize{10}{13}\selectfont
\begin{tabular}{|c|p{10cm}|}
\hline
\textbf{Tasks} & \textbf{Prompts} \\
\hline
\textbf{\makecell{Literal Description \\ 
\& Contradiction}} & \textit{The given comic with two panels shows the same situation from two opposite sides with contradictions. You need to first read and understand the comic. Generate a detailed description to illustrate the narrative of the comic and explain the contradiction of what makes the comic interesting or sarcastic.} \\
\midrule
\textbf{Underlying Philosophy} & \textit{Write a brief description of the underlying moral of the narrative in one sentence, and include what phenomenon is it satirizing and what we can learn from the comic.} \\
\midrule
\textbf{Title} & \textit{Produce a short eye-catching title reflecting the narrative.} \\
\midrule
\textbf{Negative Philosophy} & \textit{Generate five contextualized, plausible, but ultimately incorrect criticisms and moral lessons we can learn from the image, each in one sentence as distracters. Keep the length and style the same as the correct one.} \\
\midrule
\textbf{Negative Title} & \textit{Provide five seemingly reasonable, eye-catching but incorrect titles.}\\
\bottomrule
\end{tabular}
}
\vspace{2mm}
\caption{Prompts for data annotation}
\label{tab:annotation_prompts}
\end{table}

\begin{figure*}[t]
    \centering
    \includegraphics[scale=0.45]{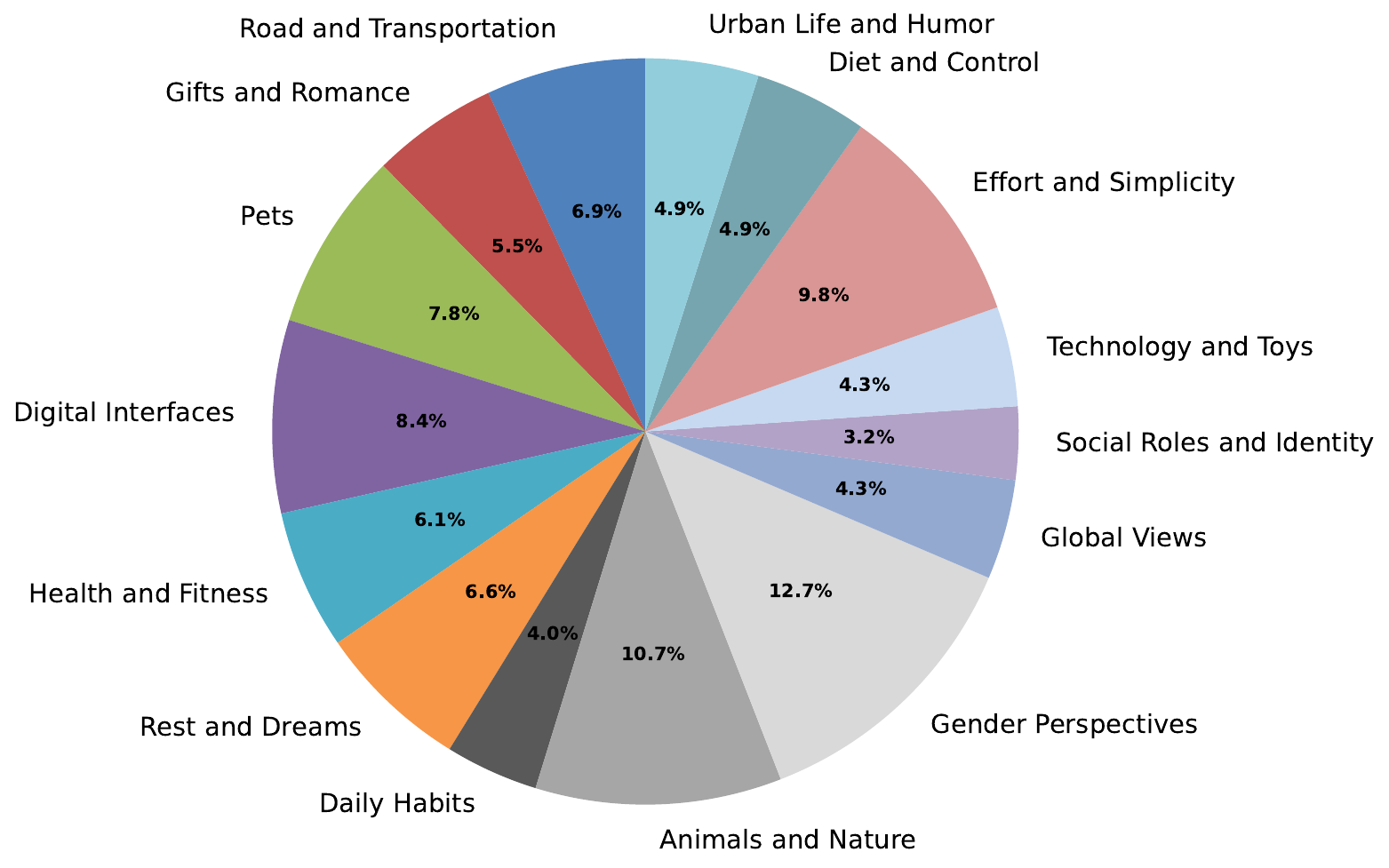}
    \vspace{-2mm}
    \captionof{figure}{The clusters of comic topics covered by our benchmark.
    }
    \vspace{-2mm}
    \label{fig:diversity}
\end{figure*}

\smallskip
\noindent\textbf{Analysis on Data Diversity.}
In order to show the diversity of our benchmark, we prompt ChatGPT to generate topical keywords for each comic based on its description, and then cluster these keywords. All these scenarios are presented in Figure~\ref{fig:diversity}. As we can see, the comics in our benchmark encompass a diverse range of everyday life scenarios. 

\section{Experimental Details}
\label{sec:exp_detail_appendix}
\subsection{Model Details}
We include both commercial and open-sourced VLMs and LLMs in our experiments. For GPT-4, we use \textit{gpt-4-vision-preview} version, and for ChatGPT we employ the \textit{gpt-3.5-turbo-0125} model variant~\footnote{\url{https://platform.openai.com/docs/models/}}. For Claude-3, we leverage the Claude 3 Ops model updated on 29th Feb, 2024~\footnote{\url{https://www.anthropic.com/api}}. For open-sourced models, we include LLaVa1.6 (34B, 13B and 7B)~\cite{liu2024llavanext}, LLaVa1.5 13B~\cite{liu2023improved}, InstructBlip 13B and 7B variants~\cite{dai2024instructblip}, CogVLM~\cite{wang2023cogvlm}, Qwen-VL~\cite{Qwen-VL}, and mPLUG-Owl2~\cite{ye2023mplugowl2}. For LLMs, we use the Llama3 instruction variant~\cite{llama3modelcard}, and the Mistral 7B instruction model~\cite{jiang2023mistral}. 

\subsection{Implementation Details}
All commercial models are accessed through their official API. For open-sourced models, we implement the experiments using Hugging Face Transformers~\footnote{\url{https://huggingface.co/docs/transformers/en/index}}. For GPT-4, Claude-3 and ChatGPT, we setting temperature as 1.0. For other models, we apply the default parameter setting or greedy decoding during inference. The experiments are conducted on NVIDIA 4090 and A6000 GPUs. 

For MCQ evaluation, we explicitly instruct the model to directly output the option in prompts, and use hard rules to parse the answer. If none of the options can be parsed, we will assign it a random option. For generation task evaluation, we apply rouge-score~\footnote{\url{https://pypi.org/project/rouge-score/}} to compute ROUGE score, and calculate the BERT score using the official implementation~\footnote{\url{https://github.com/Tiiiger/bert_score}}. For GPT based evaluations for literal description and contradiction, we use gpt-3.5-turbo-0125 version.
The prompts we used are shown in Figure~\ref{fig:prompt_gpt_eval}.

\begin{figure}[ht]
    \def\arraystretch{1.5}
	\fontsize{9}{13}\selectfont
     \hspace{-2mm}
	\setlength{\tabcolsep}{0.8mm}
	\centering
	\begin{tabular}{|p{130mm}|}
	\toprule

\textbf{Prompts for Literal Description:}\\
\hline
- Candidate literal description: {gen}

- Reference literal description: {ref}\\

Task: You need to determine how accurately the above candidate literal description matches the given reference literal description of a comic narrative.\\

Using a scale from 1 to 5, rate the accuracy with which the candidate description matches the reference description, with 1 being the least accurate and 5 being the most accurate.\\
Please directly output a score by strictly following this format: [[score]], for example: Rating: [[3]].
\\  
\midrule
\midrule

\textbf{Prompts for Contradiction:}\\
\hline
Background: You are an impartial judge. You will be given a literal description of a comic that presents the same situation from two opposing perspectives, highlighting contradictions. You will also be provided with a gold-standard illustration as reference that effectively demonstrates these narrative contradictions.\\

Your task is to evaluate the quality of a generated illustration and determine whether it accurately depicts the narrative contradictions in the comic. Then, assign a score on a scale of 1 to 5, where 1 is the lowest and 5 is the highest, based on its quality.\\

- The literal description of the comic:{description}

- The reference contradiction illustration:{ref}

- The generated contradiction illustration:{gen}\\

Please directly output a score by strictly following this format: [[score]], for example: Rating: [[3]].\\
\bottomrule
	\end{tabular}
\caption{Prompts for GPT based evaluations. 
} 
\label{fig:prompt_gpt_eval}
\vspace{2mm}
\end{figure}

\subsection{Experiment Prompts}
To reduce the biases from different prompts, we design three different prompts by different people to evaluate models and report the average results for all tasks. We present the prompts used for Literal Description (Figure~\ref{fig:prompt_exp_literal}), Contradiction Generation (Figure~\ref{fig:prompt_exp_contradiction}), Underlying Philosophy Selection (Figure~\ref{fig:prompt_exp_moral}), and Title Matching (Figure~\ref{fig:prompt_exp_title}).

\subsection{Human Evaluation Details}
\label{sec:human_eval_details}
We present 30 random samples on each task for human evaluation. We anonymize the models
and shuffle the outputs to the annotators. Following ~\cite{hwang-shwartz-2023-memecap}, we include the following aspects:

\begin{itemize}[noitemsep,nolistsep,wide]
    \item {\textbf{Correctness}}: 
    Does the model output correctly convey the narrative of the comic?
    \item {\bf Completeness:} Does the model output cover all the important elements
of the comic narrative?
    \item {\bf Faithfulness:} Can all contents from the model output be supported by the comic image (i.e., there are no hallucinations)?
\end{itemize}

For Literal description, we evaluate on all three aspects. For contradiction, we evaluate on Correctness and Faithfulness.

\section{More Sample Outputs}
\label{sec:sample_outputs_appendix}

\begin{figure*}[htbp!]
    \centering
    \includegraphics[scale=0.66]{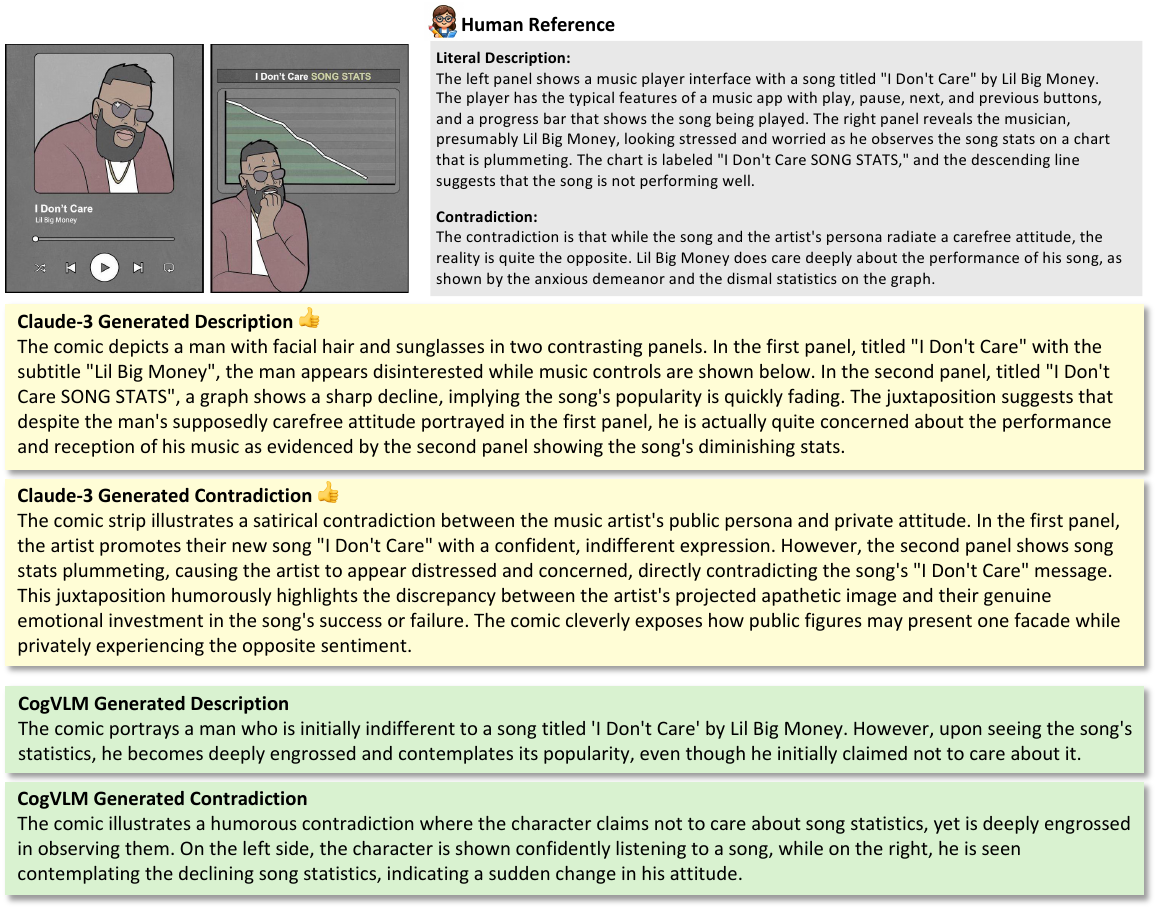}
    \vspace{-2mm}
    \captionof{figure}{Sample outputs of model generated literal description and contradiction.
    }
    \vspace{15mm}
    \label{fig:sample_output_1}
\end{figure*}

\begin{figure*}[htbp!]
    \centering
    \includegraphics[scale=0.66]{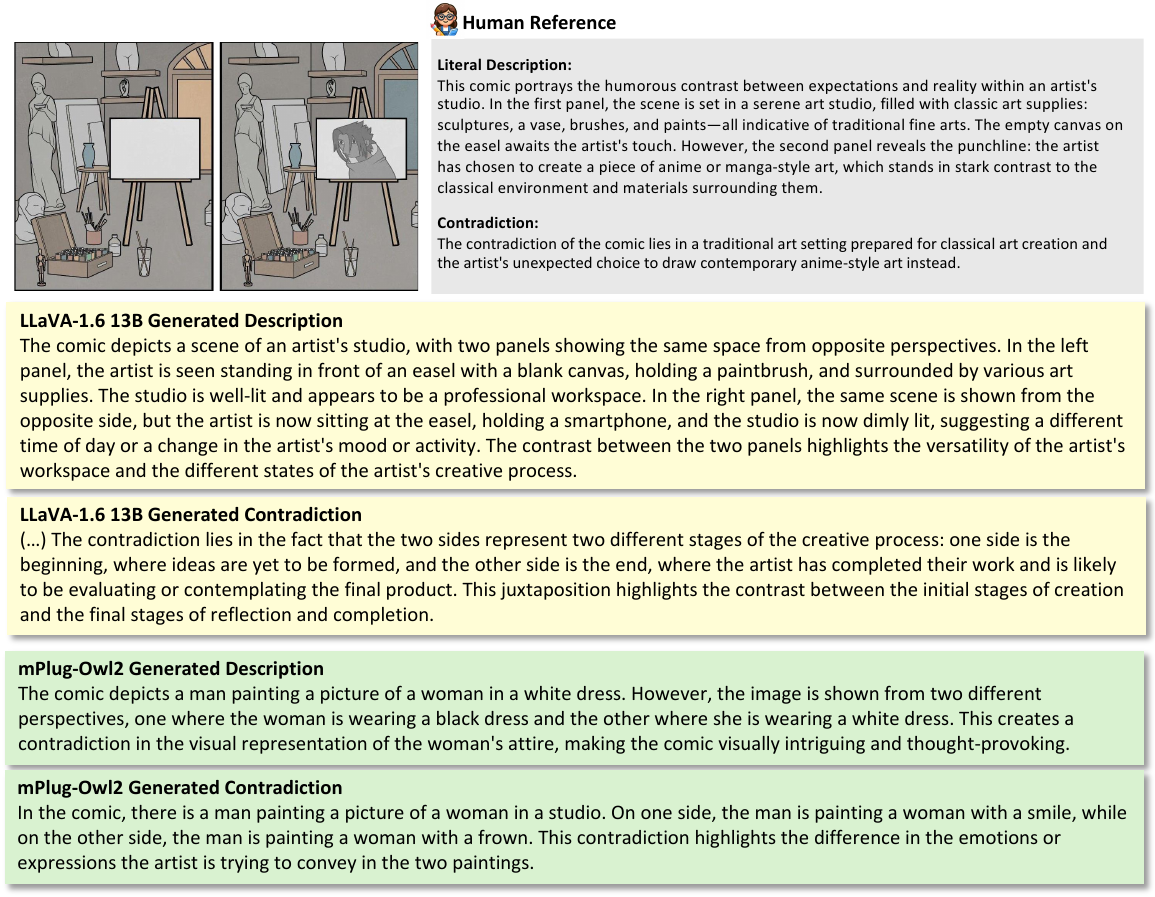}
    \vspace{-2mm}
    \captionof{figure}{Sample outputs of model generated literal description and contradiction.
    }
    \vspace{-5mm}
    \label{fig:sample_output_2}
\end{figure*}

\begin{figure*}[htbp!]
    \centering
    \includegraphics[scale=0.66]{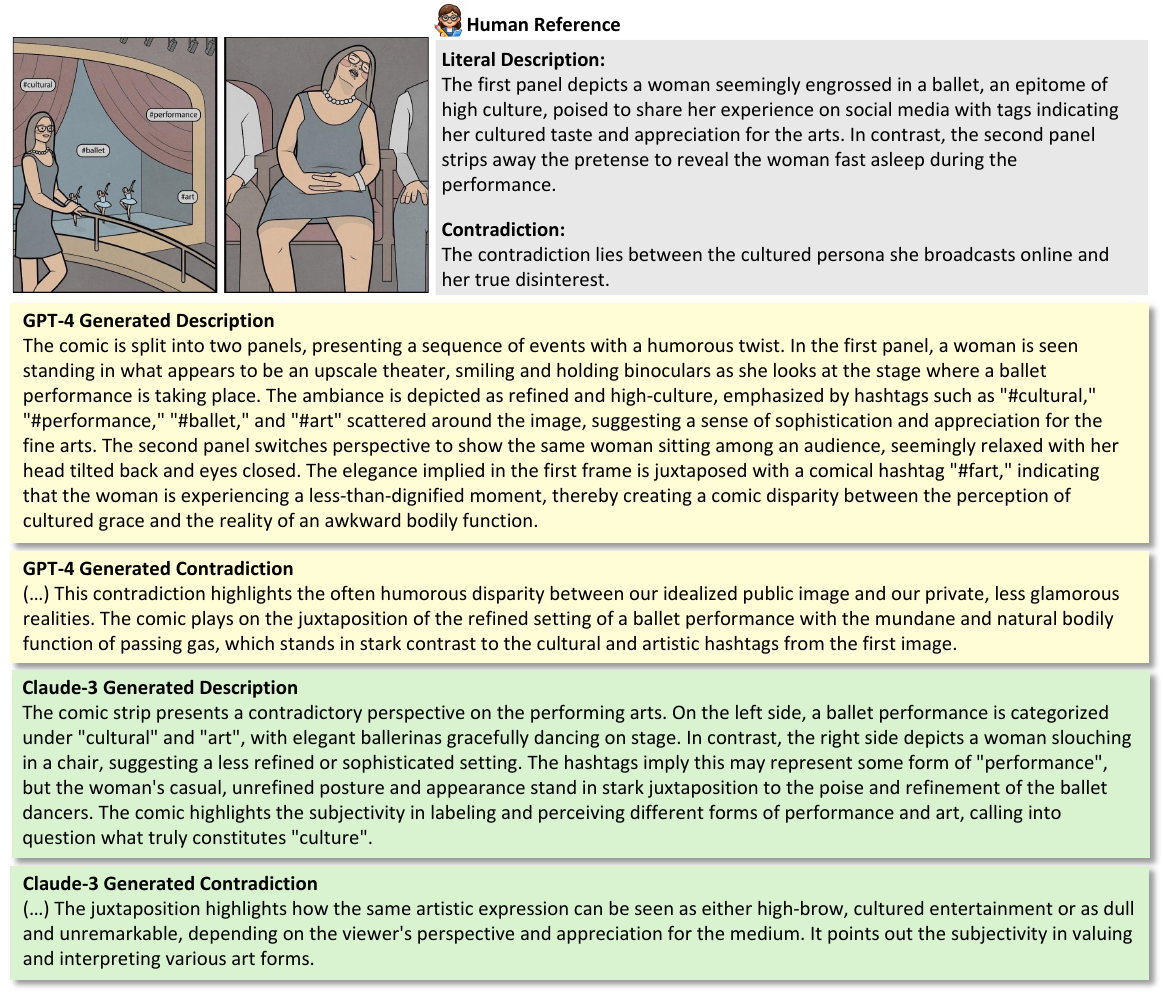}
    \vspace{-2mm}
    \captionof{figure}{Sample outputs of model generated literal description and contradiction.
    }
    \vspace{-5mm}
    \label{fig:sample_output_3}
\end{figure*}

Here, we present more randomly picked sample outputs on literal description and contradiction generation in Figure~\ref{fig:sample_output_1}, Figure~\ref{fig:sample_output_2}, and Figure~\ref{fig:sample_output_3}.

\begin{figure}[htbp!]
    \def\arraystretch{1.5}
	\fontsize{9}{10}\selectfont
     \hspace{-2mm}
	\setlength{\tabcolsep}{0.8mm}
	\centering
	\begin{tabular}{|p{130mm}|}
	\toprule
\textbf{Prompt1:}\\
\hdashline
The given comic shows the same situation from two opposite sides with contradictions. Write a one-paragraph literal description to describe the narrative of the comic.\\
\textbf{Prompt2:}\\
\hdashline
Please literally describe the context of the image in detail.\\
\textbf{Prompt3:}\\
\hdashline
Give me a detailed literal description of the image.\\
\bottomrule
	\end{tabular}
\caption{Prompts for Literal Description Generation in experiments. 
} 
\label{fig:prompt_exp_literal}
\vspace{2mm}
\end{figure}

\begin{figure}[htbp!]
    \def\arraystretch{1.5}
	\fontsize{9}{10}\selectfont
     \hspace{-2mm}
	\setlength{\tabcolsep}{0.8mm}
	\centering
	\begin{tabular}{|p{130mm}|}
	\toprule

\textbf{Prompt1:}\\
\hdashline
The given comic shows the same situation from two opposite sides with contradictions. Write a short explanation to illustrate the contradiction of the two sides.\\



\textbf{Prompt2:}\\
\hdashline

Analyze the provided image, which is divided into two or more panels, each illustrating contrasting views of the same scenario. Describe the elements visible in each panel. Then concisely interpret how these elements convey contrasting perspectives in one or two sentences. Focus and only output the contradiction.\\

\textbf{Prompt3:}\\
\hdashline
Given an image, the image is divided into two or more panels. There is the contrast relationship in the image through panels. Describe the elements visible in each panel. Give me the concise interpretation how these panels convey contrasting perspectives, which you only need to output the contradiction in one or two sentences.\\
\bottomrule
	\end{tabular}
\caption{Prompts for Contradiction Generation in experiments. 
} 
\label{fig:prompt_exp_contradiction}
\vspace{2mm}
\end{figure}

\begin{figure}[htbp!]
    \def\arraystretch{1.5}
	\fontsize{9}{10}\selectfont
     \hspace{-2mm}
	\setlength{\tabcolsep}{0.8mm}
	\centering
	\begin{tabular}{|p{130mm}|}
	\toprule

\textbf{Prompt1:}\\
\hdashline
The given comic shows the same situation from two opposite sides with contradictions.
Which of the following options best represents the underlying philosophy of the comic?

\{MCQ Options\}

Just output the choice:\\

\textbf{Prompt2:}\\
\hdashline
You are presented with an image, which is divided into two or more panels, each illustrating contrasting views of the same scenario.

Which of the following options best represents the philosophy of the image provided? 

\{MCQ Options\}

Select the correct option by typing the corresponding letter (A, B, C, or D).
\\
\textbf{Prompt3:}\\
\hdashline
Given an image, which has two or more panels. There is contrast in these panels.

Tell me the best option in the following options who represents the deep semantic of the image?

\{MCQ Options\}

Just tell me the correct option by outputing corresponding letter (A, B, C, or D), no more explanation.
\\
\bottomrule
	\end{tabular}
\caption{Prompts for Underlying Selection Task in experiments. 
} 
\label{fig:prompt_exp_moral}
\vspace{2mm}
\end{figure}

\begin{figure}[htbp!]
    \def\arraystretch{1.5}
	\fontsize{9}{10}\selectfont
     \hspace{-2mm}
	\setlength{\tabcolsep}{0.8mm}
	\centering
	\begin{tabular}{|p{130mm}|}
	\toprule

\textbf{Prompt1:}\\
\hdashline
The given comic shows the same situation from two opposite sides with contradictions.
Which of the following titles are the most suitable for the comic?

\{MCQ Options\}

Just output the choice:
\\

\textbf{Prompt2:}\\
\hdashline
You are presented with an image, which is divided into two or more panels, each illustrating contrasting views of the same scenario. Which of the following title options best represents the image provided? 

\{MCQ Options\}

Select the correct option by typing the corresponding letter (A, B, C, or D).
\\
\textbf{Prompt3:}\\
\hdashline
Given an image, the image is divided into two or more panels. There is the contrast relationship in the image through panels. 

Tell me the best title in the following title options who represents the image? 

\{MCQ Options\}

Just tell me the correct option by outputing corresponding letter (A, B, C, or D), no more explanation.
\\
\bottomrule
	\end{tabular}
\caption{Prompts for Title Matching Task in experiments. 
} 
\label{fig:prompt_exp_title}
\vspace{2mm}
\end{figure}




\end{document}